\documentclass[11pt,a4paper]{article}
\usepackage[hyperref]{ranlp2023}
\usepackage{times}
\usepackage{latexsym}
\usepackage{graphicx}
\usepackage{multirow}

\usepackage{microtype}

\usepackage{caption}
\usepackage{subcaption}
\aclfinalcopy 
\def\aclpaperid{83} 

\title{Classifying COVID-19 Vaccine Narratives}

\author{Yue Li, Carolina Scarton, Xingyi Song {\normalfont and} Kalina Bontcheva \\
  Department of Computer Science, University of Sheffield (UK)\\
  \texttt{\{yli381, c.scarton, x.song, k.bontcheva\}@sheffield.ac.uk} \\}

\date{}

\begin{document}
\maketitle
\begin{abstract}
Vaccine hesitancy is widespread, despite the government's information campaigns and the efforts of the World Health Organisation (WHO). Categorising the topics within vaccine-related narratives is crucial to understand the concerns expressed in discussions and identify the specific issues that contribute to vaccine hesitancy.  

This paper addresses the need for monitoring and analysing vaccine narratives online by introducing a novel vaccine narrative classification task, which categorises COVID-19 vaccine claims into one of seven categories. Following a data augmentation approach, we first construct a novel dataset for this new classification task, focusing on the minority classes. We also make use of fact-checker annotated data. The paper also presents a neural vaccine narrative classifier that achieves an accuracy of 84\% under cross-validation. The classifier is publicly available for researchers and journalists.
\end{abstract}

\section{Introduction}

Vaccination is one of the most effective public health interventions, but it is essential that immunisation programs are able to achieve and sustain high vaccine uptake rates. Overcoming vaccine hesitancy, which refers to the delay in the uptake or refusal of vaccines, is a major challenge \citep{eskola2015deal} and the WHO has named it one of the top ten threats to global health in 2019 \citep{qayum2019top}. Vaccine hesitancy is a complex and context specific phenomenon, varying across time, place and even vaccines \citep{larson2014understanding}. It could be caused by various factors such as concerns about side effects, costs, and misinformation. 


Although social media platforms like Twitter, Facebook, and YouTube have taken actions to limit the spread of misinformation, simply identifying and removing misinformation from platforms is not enough, as the concerns of the vaccine-hesitant citizens also need to be monitored and responded to. Consequently, fact-checkers and other professionals need analytical tools that help them to better monitor misinformation, vaccine hesitancy, vaccine-related debates and their narratives. 

Topic analysis of narratives about vaccines could be used for this purpose, however, a large manual effort is required, due to the lack of a vaccine-related topic classifier. For example, \citet{smith2020under} gather over 14 million vaccine-related posts from Twitter, Instagram, and Facebook to research vaccine-related narratives. The posts are categorised into six topics based on a novel typology designed to capture the ways narratives are framed. However, manual analysis was feasible on only a small sample of 1,200 posts, which, given the small scales, leaves significant gaps in the understanding and tackling of vaccine hesitancy.

Guided by these needs, the novel contributions of this paper are in: 

\begin{enumerate}
\item {\bf Proposing a new seven-way classification task and dataset} for categorising vaccine related online narratives. The classification task adopts the six categories (see Table~\ref{tab:CateExample}) defined by \citet{smith2020under}. The dataset is built based on manual annotation and data augmentation \footnote{We release the newly collected Twitter data: \url{doi:10.5281/zenodo.8192131}}. Our experiment demonstrates that the augmented data significantly boosts classifier performance. 
    
\item {\bf Building and making available a vaccine narrative classifier}, based on the Classification Aware Neural Topic Model (\texttt{CANTM})\citep{song2021classification}. \texttt{CANTM} originally achieved state-of-the-art performance in COVID-19 misinformation  classification \citep{song2021classification} and is particularly suited to vaccine narrative classification too, as it is robust on small training sets. For reproducibility, the classifier is publicly available as a web service \footnote{\url{https://cloud.gate.ac.uk/shopfront/displayItem/covid19-vaccine}}.
\end{enumerate}

\begin{table*}[h!]
\centering
\scalebox{0.68}{
      \begin{tabular}{|p{3cm}|p{8cm}|p{11cm}|}
        \hline
           Topic & Description & Examples\\ \hline
          Conspiracy (\texttt{Cons}) & Known or novel conspiracies and conspiracy theories involving vaccines or their development & Bill Gates: We need to depopulate the planet. Also Bill Gates: Save your life with my vaccine.\\ \hline
           Development, Provision and Access (\texttt{DPA}) & The ongoing progress or challenges concerning the development, testing and provision of vaccines as well as the access to vaccines & Oxford coronavirus vaccine triggers immune response.\\ \hline
           Liberty/Freedom (\texttt{LF}) & Civil liberties and personal freedom considerations surrounding vaccines and vaccination policies & States have authority to fine or jail people who refuse coronavirus vaccine, attorney says.\\ \hline
           Morality, Religiosity and Ethics (\texttt{MRE}) & Moral, ethical and religious concerns
           around vaccines & Kanye West Praises Trump, Hammers Planned Parenthood, Likens COVID Vaccine To ‘Mark Of The Beast’.\\ \hline
           Politics and Economics (\texttt{PE}) & Political, economic or business developments related to vaccines & Scientists Worry About Political Influence Over Coronavirus Vaccine Project.\\ \hline
           Safety, Efficacy and Necessity (\texttt{SEN}) & Safety and efficacy of vaccines, including the perceived necessity of vaccines & WHO warns coronavirus vaccine alone won't end pandemic: 'We cannot go back to the way things were'.\\ \hline
      \end{tabular}}
\caption{Description and examples of each topic.}
\label{tab:CateExample}

\end{table*}

\section{Related Work}

Since the outbreak of the COVID-19 pandemic and accompanying infodemic, large-scale monolingual and multilingual datasets have been collected from different social media platforms in order to intervene and combat the spreading of COVID-19-related disinformation \citep{shuja2021covid,alam2021fighting,shahi2020fakecovid,li2020toward,zarei2020first}, with vaccines being a commonly included topic in these datasets. As the importance of understanding and tackling COVID-19 vaccination hesitancy grew, increasing efforts have been made to analyse vaccine narratives and discourses, the dissemination of false claims and the anti-vaccine groups on social media. This has resulted in the construction of a number of COVID-19 vaccine-focused datasets, without \citep{deverna2021covaxxy,muric2021covid} or with annotations about veracity (e.g., true or false information) \citep{hayawi2022anti}, sentiment (e.g., positive, negative or neutral) \citep{kunneman2020monitoring}, stance (e.g., pro- or anti-vaccine) \citep{mu2023vaxxhesitancy,agerri2021vaxxstance,argyris2021using} or topic category (e.g., vaccine development or side effects) \citep{bonnevie2021quantifying,smith2020under}. The datasets, consequently, can be used to facilitate the research on COVID-19 vaccine-related online information from different aspects, including fact-checking, sentiment analysis, stance detection, and topic analysis.

Topics or themes discussed in the vaccine-related narratives and online debates are an essential dimension. State-of-the-art methods for automatic topic analysis typically fall under one of these categories: topic modelling \citep{jamison2020not,lyu2021covid,chen2021mmcovar,xue2020twitter}, clustering \citep{sharma2022covid,deverna2021covaxxy,muric2021covid,argyris2021using}, and inductive analysis \citep{bonnevie2021quantifying,smith2020under}. Topic modelling, represented by Latent Dirichlet Allocation (LDA) \citep{blei2003latent}, is the most commonly used approach at present \citep{jamison2020not,lyu2021covid,chen2021mmcovar,xue2020twitter}. Clustering methods for topic discovery have been applied to text representations \citep{sharma2022covid,smith2020under} or networks \citep{deverna2021covaxxy,muric2021covid}. For instance, K-means \citep{lloyd1982least} has been used to cluster the average word embeddings of vaccine narratives \citep{sharma2022covid} or to test a human-derived topic typology \citep{smith2020under}. After constructing a co-occurrence topic network with hashtags as nodes, the Louvain method \citep{blondel2008fast} is used to extract clustering from the graph \citep{deverna2021covaxxy,muric2021covid}. The above methods are unsupervised, resulting in no control on the model generation. Therefore, extra work is normally involved in discovering and labelling the topics. 

In contrast, inductive analysis relies on experts to analyse the raw textual data and derive topics or themes \citep{bonnevie2021quantifying,hughes2021development,smith2020under}. For instance, \citet{bonnevie2021quantifying} categorise anti-vaccine tweets into twelve conversation themes, such as \texttt{negative health impacts}, \texttt{pharmaceutical industry}  and \texttt{religion}. \citet{hughes2021development} identify twenty-two narrative tropes (e.g., \texttt{corrupt elites} and \texttt{vaccine injury}) and sixteen rhetorical strategies (e.g., \texttt{brave truthteller} and \texttt{appropriating feminism}) in anti-vaccine and COVID-denialist social media posts. 

Besides the above work specific to anti-vaccine contents, general COVID-19 vaccine narratives on social media were categorised by fact-checkers and researchers at First Draft \citep{smith2020under} as belonging to one of six topics, as shown in Table \ref{tab:CateExample}. 

A potential drawback of inductive analyses is that the amount of data that can be analysed by the human experts is significantly smaller than the volumes analysed through the automatic topic modelling and clustering methods. To overcome this problem, \citet{bonnevie2021quantifying} create a list of unique keywords for each theme during inductive analysis, which are then used to automatically categorise more posts based on keyword matching. 

In this paper, we explore machine learning and deep learning methods for automatic vaccine narrative classification according to the topics proposed by \citet{smith2020under}.

To the best of our knowledge, this is the first paper to frame online vaccine narrative categorisation as a classification task. In that respect, there are two closely relevant studies. \citet{song2021classification} collect English debunks about COVID-19 and annotate them with ten disinformation categories. They also propose a novel framework that combines classification and topic modelling. Similarly, \citet{shahi2020fakecovid} scrape multilingual COVID-19 related fact-check articles and manually classify them into eleven topics, but the models they explore are limited to veracity prediction. Both papers study disinformation regarding COVID-19, with vaccine covered as only one monolithic category (\texttt{vaccines, medical treatments, and tests} \citep{song2021classification} or \texttt{prevention \& treatments} \citep{shahi2020fakecovid}). However, our work is vaccine-focused, aiming at finer-grained, automatic categorisation of vaccine narratives.

\section{Vaccine Narrative Categorisation: Task Definition and Dataset Construction}

\subsection{Definition}

We define the COVID-19 vaccine narrative categorisation task as assigning COVID-19 vaccine-related claims to one of the six target topics identified by \citet{smith2020under}: (1) \texttt{Cons} for vaccine-related conspiracies; (2) \texttt{DPA} for development, provision, and access to vaccination; (3) \texttt{LF} for vaccine-related civil liberties and freedom of choice; (4) \texttt{MRE} for moral, religious, and ethical concerns; (5) \texttt{PE} for political, economic, or business aspects; and (6) \texttt{SEN} for safety and efficacy concerns. 

More detailed definitions and examples of the six topics are shown in Table \ref{tab:CateExample}.

In addition, we introduce a new, seventh category that encompasses claims related to animal vaccines (\texttt{AnimalVac}). The motivation is to recognise or filter out animal vaccine-related posts, which are also captured by keyword-based data collection methods that are typically used for collecting vaccine-related social media posts (e.g., using keywords such as vaccine or vaccines). 

Thus, this paper regards the vaccine narrative categorisation task as a seven-way classification problem, with six topics pertaining to COVID-19 human vaccination and one additional topic for animal vaccination.

\subsection{Dataset Construction}

\subsubsection{FD data} 

First Draft researchers and journalists (FD data) collected and manually annotated a number of posts in English with the six human vaccine related topics by  \citet{smith2020under}. Focusing on COVID-19 vaccine, the data covers general vaccine narratives, rather than only misinformation. It is gathered from multiple online platforms (news media, Twitter, Facebook, and Instagram), consisting of texts, images, and videos. 

For our experiments all duplicates were removed, together with posts having just video content, since our aim is text-based classification. Posts with images are classified on the basis of their textual content if available and the alternative/alt texts \footnote{a short written description of an image, which describes that image for accessibility reasons} accompanying the images.


Table ~\ref{tab:DataDistri} shows the topic distribution of the English FD dataset after data filtering is applied.

\begin{table*}[h!]
\centering
      \begin{tabular}{|c|c|c|c|c|c|c|c|c|c|}
        \hline
           & \texttt{Cons} & \texttt{DPA} & \texttt{LF} & \texttt{MRE} & \texttt{PE} & \texttt{SEN} & \texttt{AnimalVac}\\ \hline
        FD data & 26(6\%) & 116(27\%) & 37(9\%) & 7(2\%) & 108(25\%) & 134(31\%) & 0(0\%)\\ \hline
        Augmented & 107(13\%) & 116(14\%) & 92(12\%) & 151(19\%) & 108(13\%) & 134(17\%) & 96 (12\%)\\ \hline
      \end{tabular}
\caption{Distribution of data between classes before and after data augmentation.}
\label{tab:DataDistri}

\end{table*}

\subsubsection{Data Augmentation} 

As shown in Table ~\ref{tab:DataDistri}, the FD dataset is highly imbalanced. \texttt{Cons}, \texttt{LF}, and \texttt{MRE} are minority classes, which only contain 6\%, 9\%, and 2\% of the total posts, respectively. Besides, the FD dataset does not contact posts pertaining to animal vaccines, as these were excluded during their manual analysis.

To address these issues, we perform data augmentation, which includes the collection of new posts for the \texttt{AnimalVac} class, as well as gathering more examples for the three under-represented categories.

Using the Twitter API, we collected posts with vaccine-related hashtags such as \#covidvaccine, \#AstraZeneca, \#vaccines. These tweets are then filtered on the basis of class-specific keywords and hashtags which we identified manually for each target class. As we aim to limit the overlap between the FD dataset and our newly collected data, we derived the keywords and hashtags on the basis of the FD codebook, i.e. annotator guidelines:

\paragraph{Cons:} known conspiracy theories are considered, such as QAnon, ID2020, nanorobots insertion, new world order, and deep state. In addition, we included two other conspiracies fact-checked by the International Fact Checking Network (IFCN) \footnote{https://www.poynter.org/coronavirusfactsalliance/}, but not captured in the FD data: (a) The body can receive 5G signal after the vaccine is taken; and (b) China is collecting human DNA from all over the world through its vaccines in order to create a biological weapon. 

\paragraph{LF:} hashtags and terms regarding mandatory vaccination (e.g., \#MandatoryVaccine, \#NoJabNoPay), and concepts suggesting that mandatory vaccine programs undermine personal liberty or constitute a medical dictatorship (e.g., \#MedicalFreedom, \#InformedConsent, \#MyBodyMyChoice). 

\paragraph{MRE:} keywords about how people are being used as animals in vaccine testing (e.g., lab rats, guinea pigs), and about religion or ideological stance in opposition to vaccines (e.g., aborted fetuses, changing DNA). 

\paragraph{AnimalVac:} hashtags such as \#animalhealth, \#WorldAnimalVaccinationDay, and \#petmedicine are utilised to find the target tweets. As the number of the matched tweets is relatively small, we also collect Facebook posts to balance the dataset. They are picked out if they contain certain names of animal diseases and the word "vaccine". 

The full list of keywords and hashtags per class are shown with examples in Table \ref{tab:Keywords}. All posts matching the keywords and hashtags for each target class are then manually annotated by the authors, in order to ensure label quality. Table \ref{tab:DataDistri} also presents the new data distribution following this augmentation. The proportion of \texttt{Cons}, \texttt{LF} and \texttt{MRE} has increased to 13\%, 12\%, and 19\% respectively and 96 posts related to animal vaccines are also included.

\begin{table*}[h!]

\centering
\scalebox{0.68}{
      \begin{tabular}{|p{2.2cm}|p{9cm}|p{11cm}|}
        \hline
        Class & Keywords/hashtags & Examples\\ \hline
        \texttt{Cons} & QAnon, new world order, nano, ID2020, deep state, China weapon, China DNA, 5g & (1) Vaccination day. When the time comes, get vaccinated. No one will microchip you like a cat and 5G will not control your mind. \newline (2) Filled with nano particles to alter our DNA! The Moderna vaccine is the Gates vaccine.\\ \hline
        \texttt{LF} & \#freedom, \#liberty, \#NoVaccineForMe, \#MyBodyMyChoice, \#InformedConsent, \#MandatoryVaccine, \#MedicalFreedom, \#NoJabNoPay, medical dictatorship, mandatory & (1) Before you all start, this is NOT about Pro \#Vaccination or those against. This is about how the \#nojabnopay discriminates against free choice and the rich/poor. \newline (2) This is how I feel!!! We should have all of our rights and freedoms to choose what is best for us. \#freedom \#ourbodyourchoice \#NoVaccineForMe \#novaccinepassport.\\ \hline
        \texttt{MRE} & fetal/fetus/fetuses, Mark of the beast, guinea pig(s), lab rat(s), DNA, mRNA, medical ethics & (1) Vatican says use of Covid vaccines made from aborted fetal tissue is ethical. \newline (2) Africans let's rise up and put an end to this menace.. We are not lab rats!! We are not test tubes!! \#Nomorevaccinetesting\\ \hline
        \texttt{AnimalVac} & \#animalhealth, \#animalwelfare, \#WorldAnimalVaccinationDay, \#petmedicine, \#vetmedicine, Feline Panleukopenia, Feline Herpesvirus, Feline Calicivirus, Feline Leukaemia Virus, Canine Distemper Virus, Canine Parvovirus, Canine Adenovirus, Canine Rabies & (1) Will Your Pet Need a COVID-19 Vaccine? \#covid19 \#AnimalHealth \newline (2) Outbreaks of disease are unpredictable and can have a major financial impact on your farm business. Vaccination is a planned approach to help to protect your livestock and improve animal health \#VaccinesWork \#WorldAnimalVaccinationDay \newline \\ \hline
      \end{tabular}}
\caption{Keywords and hashtags for data augmentation.}
\label{tab:Keywords}
\end{table*}

\section{Predictive Model}
We evaluate feature-based and transformer-based models that are pre-trained with out-of-domain and in-domain data, and models that combine classification and topic modelling.




\paragraph{BOW-LR:}We train a Logistic Regression model with bag-of-words using L2 regularisation, using the scikit-learn implementation \citep{scikit-learn}. 

\paragraph{SCHOLAR:} \citep{card-etal-2018-neural} Sparse Contextual Hidden and Observed Language
AutoencodeR adopts VAE and directly inserts label information in the encoder during training in order to generate latent variables dependent on the labels. Zero vectors are used to represent the labels in the test set during inference. We use the author's implementation of \texttt{SCHOLAR} (\url{https://github.com/dallascard/scholar}). 

\paragraph{CANTM and CANTM-COVID} \citep{song2021classification}: Classification-Aware Neural Topic Model achieves state-of-the-art performance on COVID-19 disinformation categorisation \citep{song2021classification}. It overcomes the shortage of SCHOLAR that the label information is unavailable during inference by designing a stack of two classifier-aware VAEs. The input text is first encoded by a pre-trained Bidirectional Encoder Representations from Transformers (BERT) model \citep{devlin-etal-2019-bert}, and a classifier is jointly trained with one of the VAEs, whose generated latent variables is the input of this classifier. The other VAE takes input as the concatenation of the BERT representation and the predicted label of the classifier. The output of the decoders is the bag-of-words of the input text. To evaluate the benefit of pre-training with in-domain data \citep{gururangan-etal-2020-dont}, we also experiment with a new variant -- \texttt{CANTM-COVID} -- where we replace \texttt{BERT} by \texttt{COVID-Twitter-BERT} \citep{muller2020covid} that is pretrained on COVID-19 related tweets.

\paragraph{BERT and BERT-COVID} \citep{devlin-etal-2019-bert,muller2020covid}: We fine-tune \texttt{BERT} \citep{devlin-etal-2019-bert} and \texttt{COVID-Twitter-BERT} \citep{muller2020covid} model implemented on Hugging Face \citep{wolf-etal-2020-transformers} and follow the suggestion by \citet{song2021classification} to enable a fair comparison between \texttt{BERT} and \texttt{CANTM}: an additional 500 dimensional feed-forward network is built on top of \texttt{BERT} and the parameters, except for \texttt{BERT}'s last layer, are fixed during training.  



\section{Experimental Setup}
\subsection{Pre-processing and Hyperparameters}

All user mentions, URLs, hashtags (including those we use for data augmentation) and emojis are removed from the posts. We use the suggested settings from the original implementations \citep{song2021classification,scikit-learn,card-etal-2018-neural} except for the following hyperparameters. For each hyperparameter tuning experiment, we randomly designated 20\% of the data points in the training set as a development set. All possible combinations of candidate parameter values were tested and the optimal value was determined based on maximising the macro-F1 score on the development set. 

For \texttt{BERT}, \texttt{BERT-COVID}, \texttt{CANTM} and \texttt{CANTM-COVID}, the batch size is searched from $\{16,32,64\}$. Since FD data contains posts with long textual length, we experiment with three truncation strategies \citep{sun2019fine}: keep the beginning (the first 300, 400, or 512 tokens), the end (the last 300, 400, or 512 tokens) or a combination of both strategies (the first 300 and the last 212 tokens). The optimal selection in each experiment is keeping the first 400 tokens and training with batch size as 32. The same truncated texts are used for \texttt{BOW-LR} and \texttt{SCHOLAR}. For \texttt{SCHOLAR}, we set embedding dimension as 500, chosen from $\{300,400,500,600\}$. 


\subsection{Evaluation}

We compare the models based on 5-fold stratified cross validation on the augmented seven-class dataset. The average of macro-F1, accuracy and per-class F1 scores are reported.

\section{Results}


Table \ref{tab:overallPerf} presents the results of model comparison. The pre-trained transformer-based models significantly outperform \texttt{BOW-LR} and \texttt{SCHOLAR} whose model structures are much simpler. \texttt{CANTM} shows an increase in accuracy and macro-F1 scores compared with the strong baseline model \texttt{BERT}. Taking advantage of pre-training on an in-domain corpus of COVID tweets with a larger transformer model, \texttt{BERT-COVID} outperforms \texttt{CANTM}. \texttt{CANTM-COVID} further improves the performance, achieving the highest accuracy and macro-F1 scores. Models tend to perform better on the \texttt{Cons}, \texttt{LF}, \texttt{MRE} and \texttt{AnimalVac} classes. This is expected, since they consist of posts retrieved through class-associated keywords. 

\begin{table*}[h!]
\centering
\scalebox{0.7}{
      \begin{tabular}{|l|c|c|c|c|c|c|c|c|c|}
        \hline
        \multirow{2}{4em}{Model} & \multirow{2}{4em}{Macro-F1} & \multirow{2}{4em}{Accuracy} & \multicolumn{7}{c|}{F1 score} \\ \cline{4-10}
        & & & \texttt{Cons} & \texttt{DPA} & \texttt{LF} & \texttt{MRE} & \texttt{PE} & \texttt{SEN} & \texttt{AnimalVac}\\ \hline
        \texttt{BOW-LR} & 0.67 & 0.67 & 0.62 & 0.62 & 0.72 & 0.77 & 0.52 & 0.50 & 0.83\\
         \hline
        \texttt{SCHOLAR} & 0.65 & 0.66 & 0.65 & 0.56 & 0.67 & 0.88 & 0.46 & 0.43 & 0.89\\
         \hline
        \texttt{BERT} & 0.74 & 0.75 & 0.79 & 0.63 & 0.65 & 0.92 & 0.54 & 0.59 & 0.95\\ \hline
         \texttt{BERT-COVID} & 0.80 & 0.80 & 0.90 & 0.73 & 0.83 & 0.94 & 0.64 & 0.63 & \textbf{0.97} \\ \hline
        \texttt{CANTM} & 0.77 & 0.77 & 0.82 & 0.70 & 0.75 & 0.94 & 0.60 & 0.62 & 0.96\\ \hline
        \texttt{CANTM-COVID} & \textbf{0.84} & \textbf{0.84} & \textbf{0.91} & \textbf{0.77} & \textbf{0.86} & \textbf{0.96} & \textbf{0.67} & \textbf{0.72} & \textbf{0.97}  \\
        \hline
      \end{tabular}}
\caption{Results of model performance on the augmented seven-class test dataset. The best results are in bold.}
\label{tab:overallPerf}

\end{table*}

\section{Analysis}

\subsection{Evaluation of data augmentation} 

We analyse (1) whether our newly collected posts improve the performance on the minority classes in FD data; (2) whether the introduction of the \texttt{AnimalVac} class impacts the performance on the six human-related vaccine classes. 

\paragraph{Data Split} For the first purpose, we construct two training sets (\texttt{Training set(imbalanced)} and \texttt{Training set(balanced)}) and a test set (\texttt{Test set(six-class)}). The data of the six topics except for \texttt{MRE} in the FD data is randomly split in the ratio of 7:3 in the case of \texttt{Training set(imbalanced)} and \texttt{Test set(six-class)}. Since the \texttt{MRE} class only consists of seven posts in the FD dataset, we include them in the \texttt{Test set(six-class)} only. The newly collected \texttt{MRE} posts are randomly split in the same ratio as above to complete the \texttt{Training set(imbalanced)} and \texttt{Test set(six-class)}. The \texttt{Training set(balanced)} is the combination of \texttt{Training set(imbalanced)} and the rest of the new posts we collected during data augmentation. 

\noindent To contrast the performance before and after the introduction of the new category \texttt{AnimalVac}, we randomly split the data points in the \texttt{AnimalVac} class into two parts (7:3) and add them into \texttt{Training set(balanced)} and \texttt{Test set(six-class)} respectively, that is, \texttt{Training set(seven-class)} and \texttt{Test set(seven-class)}. Table \ref{tab:DataExp} presents the statistics of the training and test data. 

\begin{table}[h!]
\centering
\scalebox{0.65}{
      \begin{tabular}{|p{2.9cm}|c|c|c|c|c|c|c|}
        \hline
        Datasets & \textbf{Cons} & \texttt{DPA} & \textbf{LF} & \texttt{MRE} & \texttt{PE} & \texttt{SEN} & \textbf{AnimalVac}\\ \hline
        \texttt{Training set (imbalanced)} & 16 & 81 & 26 & 114 & 76 & 94 & 0\\
        \hline
        \texttt{Training set (balanced)} & 97 & 81 & 81 & 114 & 76 & 94 & 0\\
        \hline
        \texttt{Test set (six-class)} & 10 & 35 & 11 & 37 & 32 & 40 & 0\\
         \hline
         \hline
        \texttt{Training set (seven-class)} & 97 & 81 & 81 & 114 & 76 & 94 & 67\\
        \hline
        \texttt{Test set (seven-class)} & 10 & 35 & 11 & 37 & 32 & 40 & 29\\\hline
      \end{tabular}}
\caption{Label count of the training and test sets for the evaluation of data augmentation. The target classes are in bold.}
\label{tab:DataExp}
\end{table}

\paragraph{Experimental Setup} We use \texttt{CANTM-COVID} for this set of experiments as it is the best performing model as shown above. We run each experiment five times and report the average of macro-F1 and accuracy scores. 

\begin{figure*}
\centering
\begin{subfigure}{0.3\textwidth}
\includegraphics[width=1\linewidth]{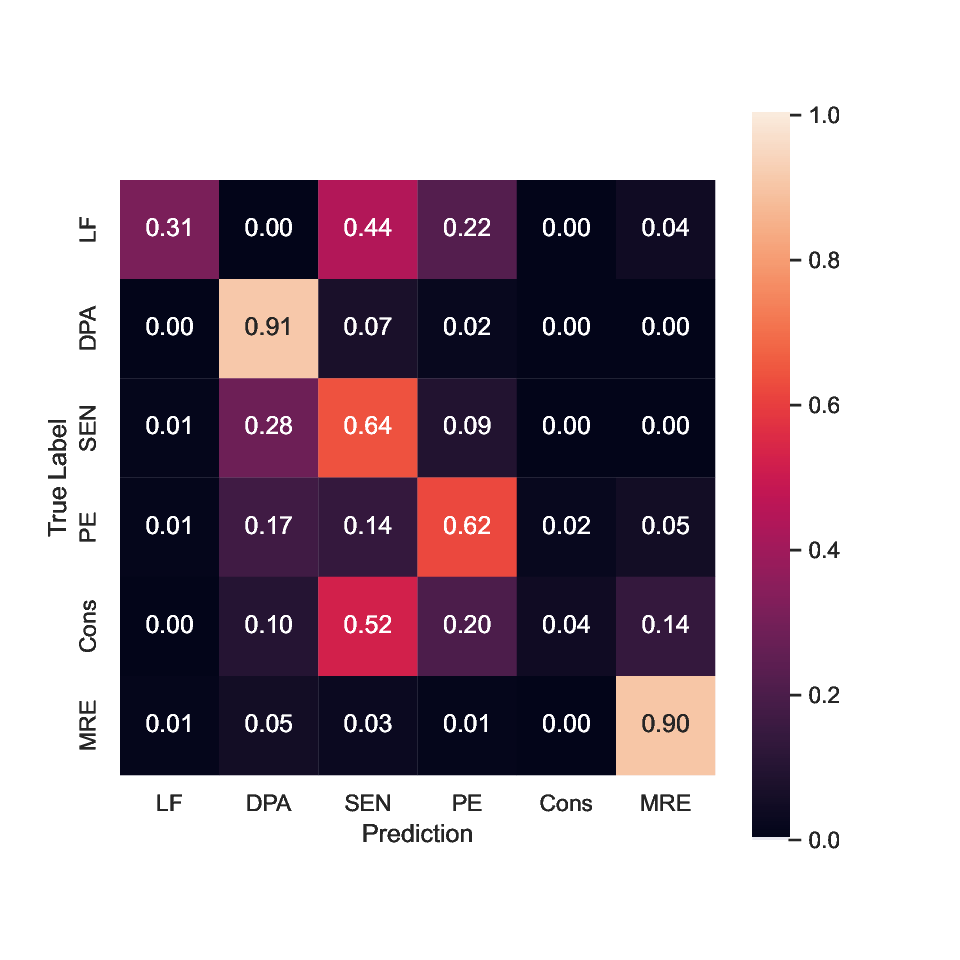}
\caption{}
\label{fig:Subfigure 1}
\end{subfigure}
\begin{subfigure}{0.3\textwidth}
\includegraphics[width=1\linewidth]{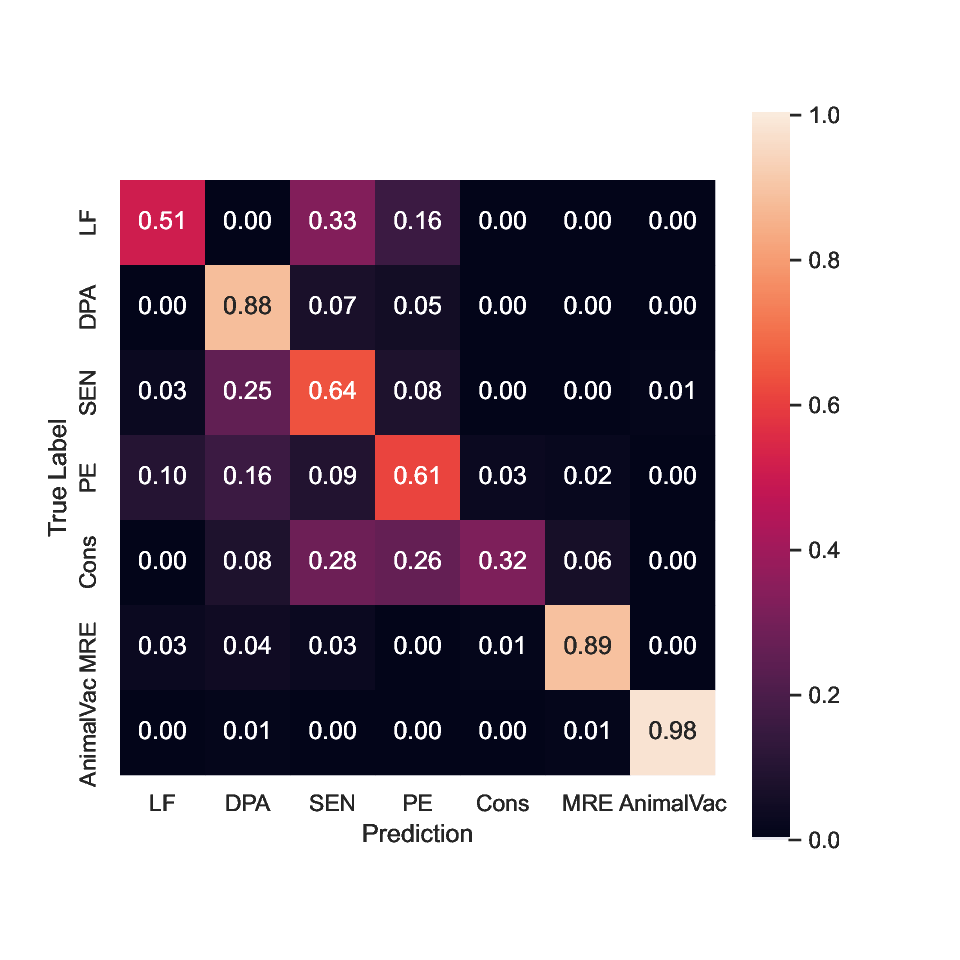}
\caption{}
\label{fig:Subfigure 2}
\end{subfigure}
\begin{subfigure}{0.3\textwidth}
\includegraphics[width=1\linewidth]{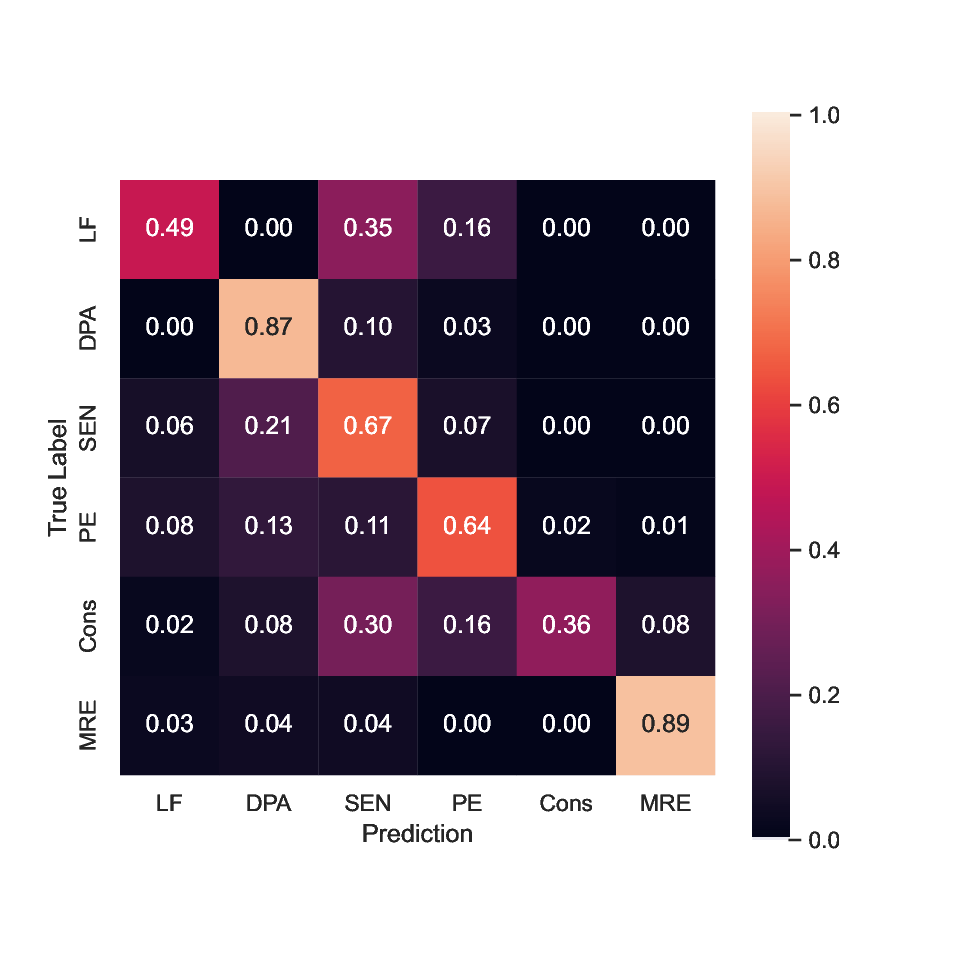}
\caption{}
\label{fig:Subfigure 3}
\end{subfigure}
\caption{Confusion matrices for data augmentation evaluation. \textbf{(a)} Model trained on six-class imbalanced data. \textbf{(b)} Model trained on six-class re-balanced data. \textbf{(c)} Model trained on seven-class re-balanced data.}
\label{Fig 2}
\end{figure*}

\paragraph{Results} The results are presented in Table \ref{tab:DataExpanResult}. We also show the confusion matrices in Fig \ref{Fig 2}. 

\begin{table}[h!]
\centering
\scalebox{0.7}{
      \begin{tabular}{|l|l|c|c|}
        \hline
        Training set & Test set & Macro-F1 & Accuracy\\ \hline
        \texttt{imbalanced} & \texttt{six-class} & 0.57 & 0.69\\ \hline
        \texttt{balanced} & \texttt{six-class} & 0.67 & 0.72\\ \hline
        \texttt{seven-class} & \texttt{seven-class} & 0.69 & 0.75\\\hline
      \end{tabular}}
\caption{Results of data augmentation evaluation of the CANTM-COVID model.}
\label{tab:DataExpanResult}
\end{table}

Re-balancing the training set could increase accuracy by 3\% and the macro-F1 score by 10\%. The recall scores of the two target minority classes (\texttt{LF} and \texttt{Cons}) grow from 0.31 to 0.49 and from 0.04 to 0.36 respectively, while the performance of the other four classes are not significantly influenced. As for the \texttt{MRE} class, 43\% of posts in FD data can be correctly predicted if training with only the newly collected tweets for this class, either in imbalanced, balanced six-class or seven-class setting. We observe that the model could accurately identify all the short tweets in \texttt{LF} after data augmentation. However, it is still hard for the model to correctly classify long posts. Details about this shortcoming are discussed in the next section. 

Introducing the \texttt{AnimalVac} class does not strongly impact the performance on the other six categories about human vaccination, which are the more important classes for this task. The model could accurately recognise 98\% of posts regarding animal vaccination, denoting that animal vaccine posts are easily distinguishable.  

As shown in Fig \ref{Fig 2}, \texttt{PE} and \texttt{SEN} posts are easily mis-classified as \texttt{DPA} (16\% and 25\% respectively. It is also hard for the model to distinguish \texttt{LF} from \texttt{SEN} and \texttt{PE}. The model struggles most on classifying the narratives about conspiracies. Only 32\% of them can be correctly tagged even after data augmentation. We discuss the potential reasons and provide examples in the next section. 

Furthermore, the drop in performance as compared to the results in Table \ref{tab:overallPerf} indicates that it is relatively easier for the model to learn and identify the augmented data collected through class-associated keyword matching, but hard to generalise to unseen domains, especially for the \texttt{Cons} class. It should be noted that we intentionally involve conspiracy stories that are not in the FD dataset (only “nano” and "deep state" appear in one post respectively after pre-processing). The \texttt{LF} class is less impacted since 95\% of new posts are collected through hashtags which are removed before training. However, our results still illustrate promising improvement in performance over the target topics, showing the ability of model generalisation.

\subsection{Error Analysis}
Although our model performs well, we highlight the following challenges and limitations. We provide some error analysis examples in Table \ref{tab:FalseExample}. 

\begin{table*}[h!]
\centering
\scalebox{0.7}{
      \begin{tabular}{|p{0.3cm}|p{1.5cm}|p{1.5cm}|p{18.0cm}|}
        \hline
        &True label &Prediction & Narrative\\ \hline
        1&\texttt{LF} & \texttt{SEN} & ....This is XXX - three months old, five days after a round of vaccines, showing the distinct sign of stroke. She died two days later....this type of asymmetry was common in the faces of the kids the day following vaccinations....Keep your eye, your focus on the MAIN GOAL: NO MANDATES period. No Mandates. No Mandates. Censorship is real.\\ \hline
        2&\texttt{LF} & \texttt{PE} & Happy to be here after spending years suffering from Trump delusion syndrome....It seems the only policy Biden has spoken about is how he will mandate masks, which ultimately will lead to vaccine mandates. Biden is in the dark in terms of medical freedom. Trump for sure.\\ \hline
        3&\texttt{Cons} & \texttt{PE} & We need to depopulate the planet. Also Bill Gates: Save your life with my vaccine.\\ \hline
        4&\texttt{SEN} & \texttt{DPA} & Good News on Covid 19 vaccine: The result of the phase two trial of the Covid 19 vaccine by Oxford University's Jenner Institute and Oxford vaccine group is very positive. The result showed a strong immune response in both parts of the immune system. The vaccine provoked a T cell response within 14 days of vaccination that can attack cells infected with the Covid 19. Participants who received the vaccine also had detectable neutralising antibodies important for protection against Covid 19. Oh God, please make this vaccine work so that we can go back to our normal world. Amen/Ameen. \\ \hline
      \end{tabular}}
\caption{Misclassification examples.}
\label{tab:FalseExample}
\end{table*}


    \paragraph{Text Length:} Long narratives involving multiple topics are easily misclassified. As shown in Table~\ref{tab:FalseExample}, the first post cites safety considerations and side effects of vaccination as grounds for objecting to mandatory vaccination. In this case, the classifier incorrectly assigns the \texttt{SEN} label. The fourth claim shows another example whose true label is \texttt{SEN} while the model falsely tags it as \texttt{DPA}. The classifier is confused because the post elaborates on the development of the COVID-19 vaccine to support the opinion towards the necessity of the vaccine in the last sentence.  

    \paragraph{Temporal Drift:} Dataset and model need to be updated over time, especially for the \texttt{DPA} and \texttt{Cons} classes, since new conspiracy theories are emerging continuously. The poor performance on the \texttt{Cons} class (see Fig \ref{fig:Subfigure 2}) illustrates that the model is finding it hard to generalise to new conspiracies. Also, progress concerning development, testing and provision of COVID-19 vaccination is fast changing. The samples in the \texttt{DPA} class were collected by First Draft in 2020 and most of the posts in their dataset refer to the announcement of the registration of the world's first COVID-19 vaccine by Russia, thus lacking examples of more recent events. Consequently we observe that the model tends to infer an unexpected correlation between Russian and the \texttt{DPA} class. 

    \paragraph{Model Bias:} The size of the current dataset is still relatively small and this may result in model bias. As shown in the second example in Table \ref{tab:FalseExample}, the mention of “Biden” and “Trump” may be the reason for the misclassification as they frequently appear in posts pertaining to politics. The class-associated words generated by \texttt{CANTM-COVID} confirm our assumption: “Trump” is highly associated with the \texttt{PE} class. Similarly, Bill Gates, who is often linked to conspiracy theories, is frequently involved in narratives about economics in the training set. In fact, “Gates” is among the top 5 topics for the \texttt{PE} class, which may explain the misclassification of the 3rd conspiracy post. The class-associated keyword-based data augmentation may also make the model overly dependent on these target terms as discussed before.


\section{Conclusion}
This paper proposed a novel seven-way classification task for categorising online vaccine narratives. We augmented an existing six-class dataset semi-automatically, leading to a more balanced data distribution and the inclusion of an additional seventh category of posts related to animal vaccines. We experimented with strong baseline models and our best model \texttt{CANTM-COVID} achieves an accuracy score of 0.84 using 5-fold cross-validation. We also show that data augmentation of minority classes helps to produce better models, without significantly impacting the performance on the remaining classes. Moreover, the addition of the new animal vaccine category does not significantly influence model performance on the original six human vaccine related classes. 

In our discussion, we highlighted the main challenges of this task and the current limitations of our model. Future work will focus on addressing some of those challenges, including development of models capable of dealing with longer posts. 

Last but not least, our vaccine narratives classifier is made available through an API for reproducibility reasons. We believe this is a significant contribution towards understanding and tracking online debates around vaccine safety and hesitancy.

\section*{Acknowledgments}


This research is supported by a University of Sheffield QR SPF Grant, an EPSRC research grant (EP/W011212/1 XAIvsDisinfo: eXplainable AI Methods for Categorisation and Analysis of COVID-19 Vaccine Disinformation and Online Debates), and an European Union Horizon 2020 Project (Agreement no.871042 under the scheme “INFRAIA-01-2018-2019 – Integrating Activities for Advanced Communities”: “SoBigData++: European Integrated Infrastructure for Social Mining and Big Data Analytics” (http://www.sobigdata.eu)). We would like to thank First Draft for data and codebook sharing and valuable feedback.

\bibliographystyle{acl_natbib}
\bibliography{ranlp2023}


\end{document}